\documentclass{article} % For LaTeX2e

\usepackage{iclr2027_conference,times}

\usepackage{amsmath,amsfonts,bm}

\def\eqref#1{equation~\ref{#1}}
\def\1{\bm{1}}

\DeclareMathAlphabet{\mathsfit}{\encodingdefault}{\sfdefault}{m}{sl}
\SetMathAlphabet{\mathsfit}{bold}{\encodingdefault}{\sfdefault}{bx}{n}

\usepackage{graphicx}
\usepackage{hyperref}
\usepackage{url}
\usepackage{xspace}
\usepackage[normalem]{ulem}
\usepackage{cleveref}

\definecolor{mygreen}{RGB}{4, 135, 76}
\definecolor{myblue}{RGB}{35, 80, 153}
\definecolor{myorange}{RGB}{220, 100, 0}

\usepackage{xspace}
\newcommand{\nvidia}[0]{Nvidia-Mamba\xspace}
\newcommand{\codestral}[0]{Codestral-Mamba\xspace}

\title{Beyond State Capacity: \\ Why Retrieval Breaks in Mamba}

\author{Nir Yarden, Amit Ben-Artzy, and Roy Schwartz \\
The Hebrew University of Jerusalem\\
\texttt{\{nir.yarden,amit.benartzy,roy.schwartz1\}@mail.huji.ac.il}
}

\iclrpreprintcopy % preprint
\begin{document}

\maketitle

\begin{abstract}
Mamba models underperform Transformers on in-context retrieval. This gap is usually attributed to Mamba's fixed-size recurrent state, which loses information as the context grows. We challenge this assumption by focusing on query-first retrieval, a setup in which the query precedes the context and state capacity should not be the binding constraint. We show that even in this setting Mamba lags substantially behind Transformers, despite its selection mechanism, which was designed to distinguish relevant content from distractors. We then demonstrate that the Mamba selection gates have sufficient control over the state updates to influence the model output: intervening on specific gates alone can swing accuracy between extremes. However, as gates are input-dependent, they require the incoming token's representation to encode its relevance to the query. We therefore probe token representations across layers, and find that the relevance signal emerges only gradually with model depth, and that the gates begin to reflect it only several layers deeper. Crucially, we find that gates in retrieval-critical layers are not yet aligned with relevance. The selection mechanism is thus expressive enough to solve the task, but it commonly operates on a relevance signal that has not yet arrived. Mamba's retrieval bottleneck therefore extends beyond state capacity: even when capacity is sufficient, the delayed emergence of relevance information across depth often prevents successful retrieval.
\end{abstract}

\section{Introduction}\label{secintroduction}

The Transformer architecture \citep{Vaswani2017Transformer} has dominated the large language model landscape for nearly a decade. Concurrently, the research community has actively sought sub-quadratic alternatives with competitive performance \citep{Katharopoulos2020Linear, Gu2021ssm}, with Mamba state-space models \citep{gu2024mamba, Dao2024mamba2, lahoti2026mamba3} emerging as a leading contender. Mamba's limitations in long-context retrieval are established in the literature, with prior work largely attributing them to its fixed-size state capacity: as the state saturates, it inevitably decays and loses information \citep{jelassi2024repeat,Waleffe2024empirical, arora2024zoology,wen2025rnns}.

State saturation provides a natural explanation for Mamba’s retrieval failures when the model is forced to store the entire sequence under limited state capacity before discovering what information will be queried. Providing the query upfront removes this pressure: the model can, in principle, identify the relevant information as it processes the context and retain only the compact target information. 
Much like a human skimming a phonebook for a specific entry, a selective recurrent model should use its input-dependent gates to filter out irrelevant distractors---precisely the behavior Mamba’s selection mechanism is intended to support. Yet empirical findings show that Mamba’s retrieval accuracy remains poor even when the query is provided before the context, falling well behind Transformers \citep{Waleffe2024empirical}. Thus, although state saturation provides an important explanation for Mamba’s retrieval failures, the poor performance in this setting points to additional factors at play, motivating us to examine whether Mamba’s selection mechanism can reliably identify and retain relevant information.

\begin{figure}[h]
\includegraphics[width=\linewidth]{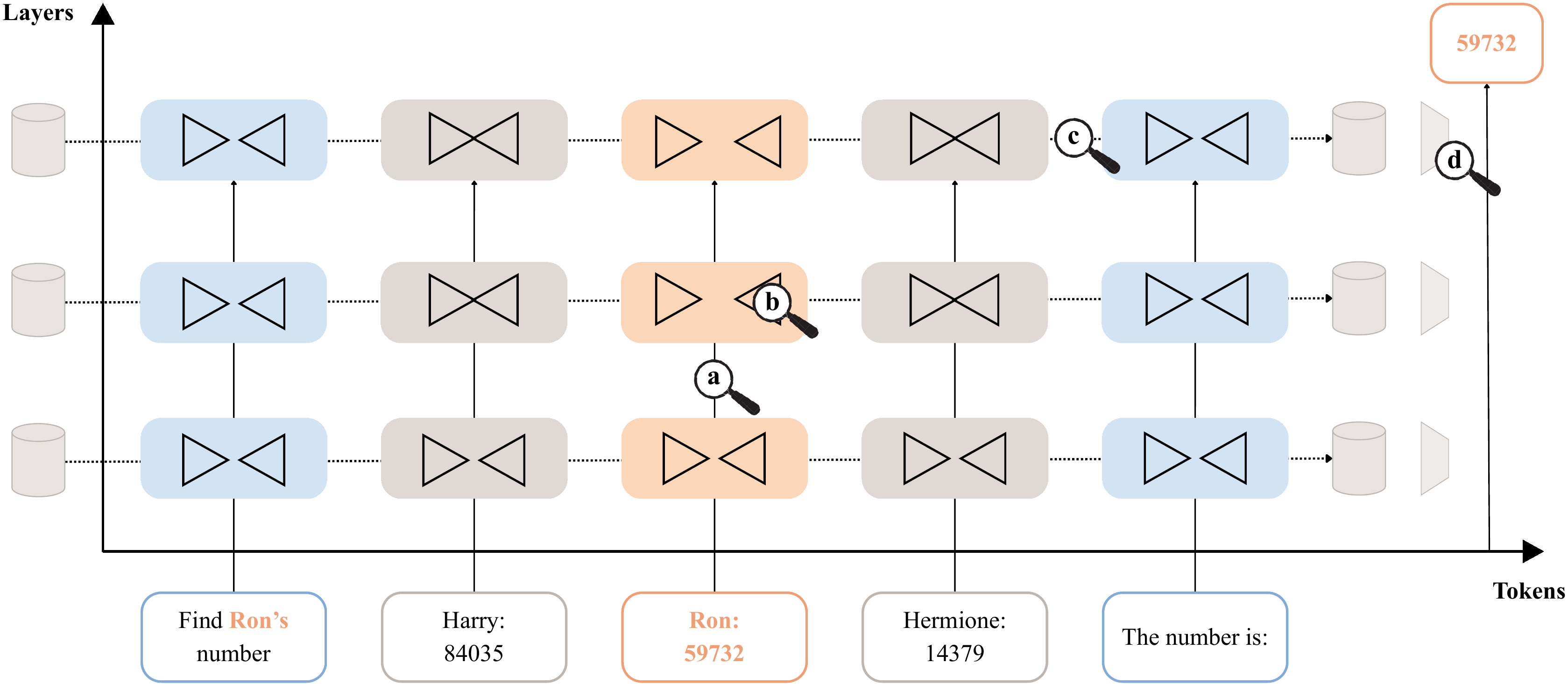}
\caption{\textbf{Mechanistic overview of query-first retrieval in Mamba}\quad Guided by the design principles behind Mamba's architecture, we trace the flow of information across the layer and token-sequence axes, examining under a magnifying glass the mechanisms involved at query-first retrieval in Mamba, in search of the source of its degraded performance. Specifically, we zoom in on the following processes: (\textbf{a}) As token representations are mixed across layers, a relevance signal is encoded. (\textbf{b}) The selection gate responds to this signal by `opening and closing', selectively writing relevant content to state while suppressing distractors. (\textbf{c}) The recurrent state stores the target information while being repeatedly updated, which requires protecting it from decay. (\textbf{d}) Finally, the target information is extracted from the state to generate the output token.}
\label{fig:figure_1}
\end{figure}

To uncover the root cause of this failure, we work backward from the output to trace where Mamba's information flow breaks down~(\cref{fig:figure_1}). First, we construct a controlled setting using a query-first key-value retrieval task---concretely, a synthetic phonebook in which the model must return the value paired with a queried key, embedded among many distractor entries. We find that simply prepending a relevance flag to the target entry drastically improves Mamba's accuracy. The flag explicitly signals the entry’s relevance instead of relying on contextual deduction. We can therefore localize the bottleneck by comparing information flow between the flagged and unflagged variations. Specifically, we track the recurrent states to determine when the target information is retained, and analyze the selection mechanism's gates ($\Delta$) to assess how effectively they distinguish target tokens from distractors. Next, we probe the target entry token's representations to measure how deeply its contextual relevance is encoded and how rapidly this signal reflects upon the selection gates. To measure the gates' exact control over performance, we patch gate values between the variants and inject oracle gate spikes at critical tokens.

Ultimately, we find that the selection gates strictly govern retrieval accuracy: trivial modifications to these gates can dramatically restore or destroy Mamba's output. This demonstrates that the selection mechanism is capable of effective state updates, provided its gates are well-calibrated. The problem, however, is that the gates are often miscalibrated: long-range contextual information requires many layers to mix into a token's representation and then reflect upon its gates. Empirically, critical state updates tend to occur before the token's relevance is adequately indicated in the selection mechanism. These findings reveal that limited state capacity alone does not fully account for Mamba's retrieval difficulty, and point to selection gate calibration---shaped by both architectural design and training---as a critical factor in addressing it.

\section{Background}\label{sec:background}

Built upon the State Space Model (SSM), each Mamba block maintains its own hidden state $h_t$ that recurs across the sequence, taking as input at each position $t$ the token representation $x_t$ which is the previous block output. Within a block, the selective state-space transition is defined as
\begin{align}
    h_t = \overline{\mathbf{A}}_t \mathbf{h}_{t-1} + \overline{\mathbf{B}}_t x_t, &\qquad y_t = \mathbf{C}_t^\top h_t \\
    \overline{\mathbf{A}}_t = e^{\Delta_t \mathbf{A}} ,&\qquad
    \overline{\mathbf{B}}_t = \Delta_t \mathbf{B}_t
\end{align}
where $\overline{\mathbf{A}}_t$, $\overline{\mathbf{B}}_t$, and $\mathbf{C}_t$ are the state decay, write, and read matrices, respectively, and the selection gate $\Delta_t$ is a token-dependent discretization timescale that gates the update of the hidden state: high $\Delta_t$ nearly resets $h_t$ to prioritize $x_t$, while low $\Delta_t$ discounts $x_t$ to preserve state continuity.

Crucially, $\Delta_t$, $\mathbf{B}_t$, and $\mathbf{C}_t$ are not learned constants but are themselves computed from the current token representation, via a \emph{selection} mechanism applied to $x_t$:
\begin{equation}
    (\Delta_t, \mathbf{B}_t, \mathbf{C}_t) = \mathrm{selection}(x_t)
\end{equation}
implemented in practice as a set of linear projections of $x_t$. This makes the state update \emph{selective}: every parameter governing how much of $x_t$ is written into $h_t$, how quickly the state decays, and how $h_t$ is read out is derived exclusively from $x_t$ itself.

The Mamba architecture \citep{gu2024mamba} is built by stacking Mamba blocks, each combining this SSM sequence mixing with MLP-like feature transformation, as well as a short convolution---the resulting block's output becomes the next block's $x_t$. This alternation between a token-mixing operation and a per-token feature transformation mirrors the structure of the standard transformer block, which stacks attention layers with feed-forward layers. Across layers, this repeated alternation acts as a pipeline that progressively enriches each token embedding with both linguistic and context-dependent semantic information \citep{bertPipline2019, mambaPipeline2025}.

\section{Methodology: Information-Flow Diagnosis}\label{sec:experiments}

\paragraph{Experimental strategy} To understand the factors underlying Mamba’s retrieval failures beyond the challenges imposed by its finite state capacity, we need a setting in which the availability of relevance information can be precisely controlled. We therefore study a query-first retrieval setting in which a query specifies a target entry in a key-value context, and construct two nearly identical task variants: one provides only the query, requiring relevance to be inferred from the broader context, while the other additionally provides an explicit local relevance flag at the target entry. Because the two variants differ only in the availability of this local relevance flag, the resulting performance gap provides a direct measure of how much Mamba benefits from having relevance information available locally, while holding the retrieval problem itself fixed.
\paragraph{Diagnostic framework} To understand the source of the query-first retrieval performance gap, we structure our analysis around the stages of processing that follow from the motivations behind Mamba’s architectural design. The selection mechanism at each layer is parameterized solely as a function of that layer’s token representation $x_t$. Thus, for selection to act on relevance, relevance must first be accumulated in the token representations through the token-mixing process across layers. We examine how this relevance emerges in the representations, and whether the selection gates respond to it by favoring target tokens over distractors. Because $\Delta_t$ jointly governs state writing and retention, favoring target tokens also requires suppressing updates at subsequent distractor tokens to preserve the target information in state—an imperative well established in recurrent architectures~\citep{Westhuizen2018, Schlag2021, yang2025}. We then consider whether the retained information can ultimately be extracted to produce the output. Across these stages, we compare the model’s processing with and without the local relevance flag to identify where its availability changes the model’s behavior.

\paragraph{Datasets}\label{para:datasets}
We construct a synthetic query-first phonebook dataset based on the \textit{reversed} setup of~\citet{Waleffe2024empirical}, itself introduced in the phonebook task of \citet{jelassi2024repeat}. Each prompt contains a natural-language description of the task and two in-context examples illustrating the phonebook format, followed by a phonebook and a request to recall the phone number associated with a target name. An example is shown in \cref{fig:prompts}. Each phonebook entry follows the format \texttt{John Doe: 3948874628\textbackslash n}. We refer to the queried phonebook entry as the \emph{target entry}, and to the phone number it contains as the \emph{target information}. We consider two variants: in the \emph{unflagged} dataset, the target entry is indistinguishable from the other entries; in the \emph{flagged} dataset, which we introduce, the target entry is explicitly marked, for example, \texttt{\textbf{target} Jane Doe: 3827954899\textbackslash n}, providing an immediate local relevance signal.

\paragraph{Models} We experiment with two Mamba-2 models: \textbf{(a)} An 8B base-model pretrained for general purpose language modeling ~(\textbf{`\nvidia'}; \citealp{Waleffe2024empirical}); and \textbf{(b)} a 7.3B instruction-tuned model with emphasis on code completion ~(\textbf{`\codestral'}; \citealp{codestral2024blog}).

Full implementation details and hyperparameters can be found in \cref{appendix:experiments_details}.

\begin{figure}[t]
\includegraphics[width=\linewidth]{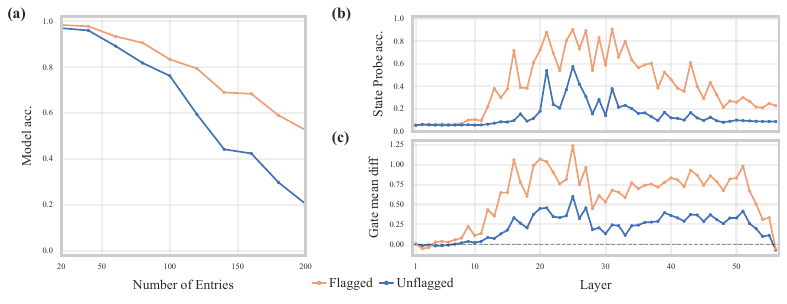}
\caption{A local relevance flag in the target entry enables Mamba to identify and encode relevant information earlier, substantially improving long-context retrieval. (\textbf{a}) Evaluating the model on the query-first phonebook dataset, with and without a local relevance flag. The flag drastically improves Mamba's accuracy on longer phonebooks. (\textbf{b}) Probing recurrent states for the target information reveals that the local relevance flag enables target decodability to emerge in earlier layers and with greater strength. (\textbf{c}) Calculating the difference between Mamba's selection gates for target and distractor tokens demonstrates that the selection gate distinguishes token relevance earlier and more effectively when the local relevance flag is included. Showing results for \nvidia; See \cref{fig:codestral_corralation} for \codestral results.}
\label{fig:nvidia_corralation}
\end{figure}

\section{Local Relevance Signals Enhance Selective State Updates}\label{sec:correlation}

\citet{Waleffe2024empirical} showed that Mamba underperforms Transformers in query-first retrieval. We demonstrate that adding a local relevance flag to the target entry substantially improves accuracy, and we analyze state updates and gating dynamics across both settings to further localize the source of this failure.

\subsection{Local Relevance Rescues Retrieval}\label{subs:evaluation}

We evaluate Mamba's accuracy in retrieving phone numbers across the two variants of the phonebook task, \emph{unflagged} and \emph{flagged}, as the number of entries in the phonebook increases. Our results reveal that the \emph{unflagged} setup exhibits a trend of gradual performance degradation as context length grows. In contrast, prepending a local flag to the target entry substantially improves accuracy. While performance still degrades with longer contexts, the decline is markedly less severe~(\cref{fig:nvidia_corralation}a).

Given that the two variants differ solely by a single word which substantially improves the performance---more than doubles \nvidia's accuracy on long phonebooks---we conclude that Mamba processes tokens more effectively when their relevance is indicated locally rather than established globally hundreds of tokens earlier.

\subsection{Target Information is Less Available in Recurrent State}\label{subs:state_probing}

Next, we show that the failure stems from improper storage of the target information within recurrent state updates rather than an inability to extract it at output time. We evaluate how well the target information is encoded in the recurrent state by capturing snapshots of the state in entries following the target, then training layer-wise probes \citep{Bengio2016probing} to predict the target information from the flattened state. To frame this as a classification task suitable for a linear probe, we modify the dataset for this specific experiment: each entry consists of a name and a profession drawn from a closed set of 20 professions, rather than a phone number. We experiment both with the \emph{unflagged} and \emph{flagged} variants.

Evaluating the probes on the test set (\cref{fig:nvidia_corralation}b) reveals that target decodability from the recurrent state is consistently higher on the \emph{flagged} dataset and emerges at considerably earlier layers compared to the \emph{unflagged} dataset. Because state decodability differences between the variants align with accuracy differences, this finding directs our remaining experiments toward the mechanisms responsible for updating the state, rather than those that read from it.

\subsection{The Selection Gating struggles to Distinguish Target from Distractors}\label{subs:gate_distinction}

We have shown that Mamba's performance improvement with a local relevance flag stems from changes in its state updates. Next, we demonstrate that this effect is also reflected in the selection gates. For this purpose, we evaluate Mamba's gating mechanism's ability to distinguish between target and distractor tokens. \Cref{fig:nvidia_corralation}c shows the difference between the average gate value assigned to target information digit tokens and distractor digit tokens across layers. The gate value, commonly denoted $d_t$, is the direct output of the selection projection for the input token\footnote{Unlike the subsequent experiments, this experiment uses the pre-activation projection output $d_t$ rather than $\Delta_t$, as averaging after a non-linear function introduces artifacts; see \cref{appendix:experiments_details} for details.}, averaged over all heads in the layer and samples in the dataset.

Evidently, Mamba's gates achieve a substantially better distinction between target and distractor tokens when the target entry includes the local relevance flag. Ultimately, lower accuracy aligns with inadequate gating in the selection mechanism.

\begin{figure}[h]
\includegraphics[width=\linewidth]{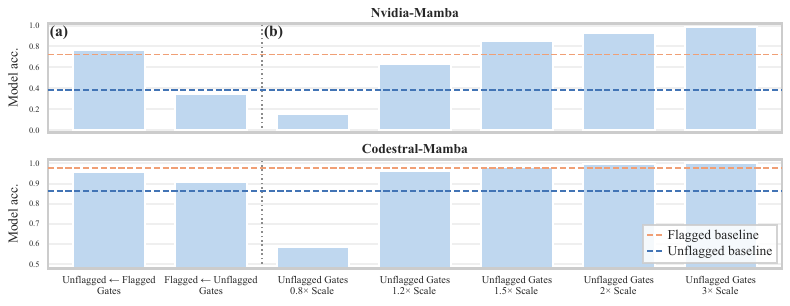}
\caption{Manipulating the gates corresponding to the target information tokens governs performance. (\textbf{a}) Patching the \emph{unflagged} prompt with \emph{flagged} gating values recovers full \emph{flagged} accuracy, whereas the reverse substitution reduces \emph{flagged} accuracy to \emph{unflagged} levels. (\textbf{b}) Oracle-scaling the target information tokens gates of the \emph{unflagged} prompt can bring Mamba to near-perfect or near-zero accuracy. }
\label{fig:causality}
\end{figure}

\section{Selection Gates Govern Retrieval Accuracy}\label{sec:causality}

We demonstrated that incorporating a local relevance flag into the target entry substantially improved retrieval accuracy. This improvement is aligned with enhanced state decodability of the target information and a sharper distinction in the selection gates between target and distractor tokens. Next, we aim to establish a causal link showing that this gating mechanism directly governs model accuracy in query-first retrieval.

\paragraph{Bidirectional patching transfers performance} We test whether the ability of the selection gating mechanism to separate target and distractor entries causally affects accuracy, beyond the aforementioned correlation. We run Mamba on each sample's \emph{unflagged} and \emph{flagged} variants to collect both the baseline accuracy and the selection gate values assigned to the target information tokens. Next, we run Mamba on the \emph{unflagged} prompt, patching \citep{wang2023patching} the gate values of its target information tokens with those captured in the baseline \emph{flagged} run. Strikingly, this targeted intervention on the selection gates of merely 10 tokens completely shifts the model's accuracy on the \emph{unflagged} dataset to nearly match its baseline performance on the \emph{flagged} dataset (\cref{fig:causality}). The reverse dynamic holds as well: when patching the target entry tokens of the \emph{flagged} prompt with gate values from the \emph{unflagged} run, Mamba drops to the lower accuracy of the \emph{unflagged} baseline.

\paragraph{Oracle scaling demonstrates sufficiency} The substantial impact of properly gating target tokens is further corroborated when, instead of patching the \emph{unflagged} gate values, we scale them by a constant factor to spike them---consequently boosting the writing of the target information into the recurrent state. \Cref{fig:causality} demonstrates that using the ground-truth indices of the target information tokens to oracle-scale their selection gates achieves nearly perfect accuracy. Conversely, dampening the same gates with a scaling factor less than 1 severely degrades performance. This indicates that well-calibrated gating is a pivotal factor in Mamba's success.

\begin{figure}[h]
\includegraphics[width=\linewidth]{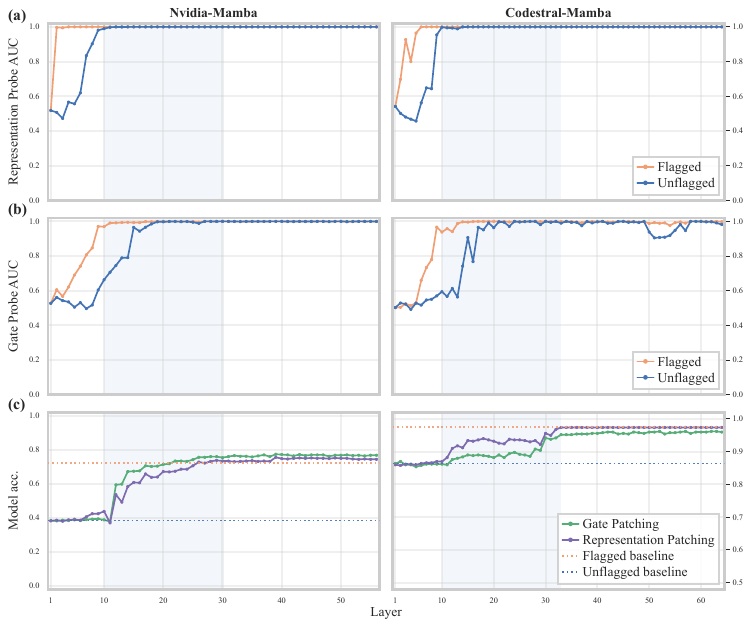}
\caption{Relevance information emerges gradually through the network, with delays both in contextualizing token representations and in aligning the gates with the resulting relevance signal. (\textbf{a}) Layer-wise linear probing shows that relevance becomes readily decodable from token representations much later when a local relevance flag is not present. (\textbf{b}) Layer-wise probes reveal a further delay: relevance becomes decodable from the gates only several layers after it emerges in the corresponding representations. (\textbf{c}) Cumulative layer-wise patching of either target information gates or representations shows that accuracy gains are concentrated in the early-to-middle layers, including layers where relevance is already decodable from representations but not yet adequately reflected in the gates.}
\label{fig:failure}
\end{figure}

\section{Relevance Signals Emerge Too Late for Effective Retrieval}\label{sec:failure}

We previously established that Mamba's gating mechanism plays a causal role in its accuracy. In this section, we show that a token's relevance signal emerges only gradually across layers, first becoming decodable in the token's representation and only later surfacing in the corresponding gates. Critically, this process is slow enough that by the time the gate reflects a token's relevance, layers that play a role in state updates critical for retrieval have often already passed.

\subsection{Relevance Emerges Late in Token Representations}\label{subs:representation_probing}

Since the selection mechanism is essentially a single linear layer, we investigate how deep into the residual stream a token's relevance signal becomes linearly decodable. For each entry's final digit token, we extract its representation at every layer and train layer-wise linear probes to classify target versus distractor tokens. We report probe performance using AUC, which quantifies the separability of the two classes.

\Cref{fig:failure}a reveals a substantial difference between the \emph{unflagged} and \emph{flagged} datasets in the number of layers required for the relevance signal to become readily decodable. When Mamba processes a \emph{flagged} sample, the relevance signal becomes linearly decodable within one to two residual-stream updates. In contrast, in the \emph{unflagged} setting this decodability only plateaus after nearly 10 layers, indicating that contextual enrichment of the representation is markedly slower when the query is separated from the target by long-range context.

\subsection{Gates Lag Behind Representations}\label{subs:gate_probing}

We next test whether relevance decodability in the representation is necessary for well-calibrated gates, and characterize the resulting dynamics. At each layer, the selection mechanism produces a gate vector with one entry per head for every token. As in the representation-probing analysis above, we extract the gate vector at each entry's final digit token and train layer-wise probes to classify target versus distractor tokens.

As shown in \cref{fig:failure}b, the gate vector requires additional layers beyond the representation itself before relevance becomes decodable. This pattern holds for both the \emph{flagged} and \emph{unflagged} datasets: gate-level decodability rises gradually over roughly 10 layers, beginning near the layer at which the corresponding representational decodability first emerges. A natural interpretation is that relevance signals emerge later in the \emph{unflagged} representations, with this delay affecting the gates, correspondingly pushing the onset of gate-level decodability.

\subsection{Critical State Updates Occur Before Gate Alignment}\label{subs:patching2}

%\roy{I am not sure I follow this experiment. Is this identical to section 5 patching, just accumulative rather than patching a single layer? why is this important here?}

We extend the gate-patching experiment from \cref{sec:causality} to measure layer-wise dynamics. While the original experiment patched gates across all $N$ layers simultaneously, we now patch an increasingly deep prefix of layers (1, 1--2, 1--3, ... , 1--$N$) and measure the resulting change in accuracy. This lets us pinpoint not just \textit{whether} gate patching causally affects accuracy, but \textit{where} in the network's depth this effect occurs---a result we can connect to our findings on relevance-signal propagation dynamics. As \cref{fig:failure}c shows, patching the gates with their \emph{flagged} counterparts yields small accuracy gains that accumulate across a contiguous range of layers as progressively deeper layers are included. Notably, this effect is concentrated in the early-to-middle portion of the network, rather than spanning the model's full depth.

We additionally repeat the experiment, this time patching the token representation rather than the selection gates. Since the representation is the input of the selection mechanism, patching it indirectly corrects $\Delta_t$, $B_t$, $C_t$, thereby affecting the entire state update and readout within the layer. The resulting accuracy pattern closely mirrors that of gate patching alone. This similarity suggests that within this critical layer range, miscalibrated gating---rather than miscalibration of the other selection components---is the dominant causal factor limiting accuracy.

Comparing across these experiments, we find that although the relevance signal is already perfectly decodable in the representation throughout this critical range, the gates fail to reflect it adequately in the range's earlier layers---a gap which is consistent with the limited \emph{unflagged} accuracy. Together, these results expose a core limitation of Mamba's query-first retrieval: when the query appears exclusively in a distant context, aligning the state-update gating mechanism with the true relevance signal requires substantially more layers of representational processing than when the query appears near the target information.

\section{Related Work}\label{sec:related_work}

\paragraph{Mamba for in-context retrieval} Our work investigates query-first retrieval, a setting in which Mamba's state-capacity limitations are insufficient to explain performance drops. Prior work has largely focused on evaluating the structural bottlenecks of recurrent architectures in scenarios where the query is placed at the end. Both \citet{jelassi2024repeat} and \citet{Waleffe2024empirical} highlighted Mamba's inferior performance relative to Transformers on copying and retrieval, identifying its fixed state size and fuzzy memory bottleneck as primary causes. Evaluating across complex associative tasks, \citet{arora2024zoology} introduced the Multi-Query Associative Recall (MQAR) benchmark, confirming that recurrent architectures struggle when multiple associative recalls are required. Formally, \citet{wen2025rnns} proved that the maximum context length a recurrent model can reliably retrieve is bounded linearly with its state size, establishing a hard theoretical limit on Mamba's retrieval capabilities.

\paragraph{SSM-Transformer hybrid models} While Mamba struggles with in-context retrieval due to its fixed-capacity nature, Transformers generally perform well on such tasks, which is commonly attributed to self-attention’s ability to directly access individual representations of prior tokens. This contrast has motivated hybrid architectures \citep{Waleffe2024empirical,Glorioso2024Zamba,Dong2024Hymba,lenz2025jamba} that interleave SSM and attention layers, using attention to provide precise token-level retrieval. These approaches have shown promising results and provide an effective way to mitigate the capacity bottleneck of pure SSMs. Our findings highlight, however, that fixed state capacity does not fully account for Mamba’s retrieval failures: even in settings where state capacity is not the limiting factor, Mamba can fail to selectively retrieve information. This suggests that improving the retrieval mechanisms of SSM layers remains valuable both for pure SSMs and for hybrid architectures, whose SSM components are still expected to efficiently process and retain relevant information.

\paragraph{Mechanics of Mamba over long contexts} Several recent works localize the
components behind Mamba's long-context failures. \citet{decimamba2025} attribute
degradation to an effective receptive field dictated by the training sequence length,
and use the selection gate as a token-importance score to compress the context.
\citet{bick2025gather} compare Transformers and SSMs and find retrieval carried by a
few specialized \emph{Gather} and \emph{Aggregate} heads, with SSMs implementing the
aggregate step less sharply. \citet{longmamba2025} separate local from global hidden
channels and trace long-context memory decay to the latter. These works ask which
component is responsible---which heads, which channels. We ask instead whether the
responsible component has what it needs at the moment it acts, locating the failure in
the timing of information within the network rather than in the identity or capacity of
any component.

\paragraph{Mechanistic interpretability methods} Our experimental framework draws upon established diagnostic probing and causal intervention frameworks. To inspect internal representations, probing classifiers \citep{Bengio2016probing} offer a standard diagnostic tool. However, in light of known limitations in probing methodologies \citep{Ravichander2020Probing, belinkov22probing}, we treat probe accuracy as correlative, recognizing that extractable information may not be causally used downstream. To establish causality, we employ patching---drawing on causal mediation analysis \citep{vig2020patching} and path patching \citep{wang2023patching}---by replacing a specific component in a forward pass with its counterpart from a minimally edited counterfactual input. Adhering to activation patching guidelines \citep{heimersheim2024patching}, we apply bidirectional interventions to confirm both the necessity and sufficiency of these mechanisms.

\section{Conclusions}\label{sec:conclusions}\

In-context retrieval is widely recognized as a foundational capability driving core LLM behavior~\citep{Olsson2022Incontext, Ram2023InContext}, and the fact that it poses a major difficulty for Mamba is a key constraint limiting its adoption. This work demonstrated that fixed state capacity is not the sole mechanistic bottleneck affecting retrieval, by tracing Mamba's information flow in query-first retrieval---a setting where large state capacity is unnecessary. By directly intervening on Mamba's selection mechanism, we found that naively adjusting the gate values of only a handful of tokens is enough to make or break retrieval accuracy, and that swapping gate values between two prompts transfers their outcomes, turning a successful retrieval into a failed one and vice versa. This shows the selection mechanism is well-suited to govern retrieval accuracy through its control over state updates, as it is sufficiently expressive to determine the outcome on its own. However, we found layers critical for retrieval in which the gates are not well calibrated. This is because the relevance signal only emerges deep into the layer stack, past the point where these gates already need to act on it.

This work leaves several important open questions. First, the observed performance gap may be rooted in standard training regimes, which often prioritize local sequence modeling over explicit global context synthesis, causing a mismatch between the context encoded in token representations and the inputs required for the selection mechanism to gate successfully. Second, and complementarily, Mamba's current design---relying exclusively on a token's isolated representation to compute its gate—may introduce a fundamental bottleneck. This forces the model to expend substantial depth and capacity simply compiling global context into single-token representations---a process we found completes too late in the network's depth to influence all decision-critical layers. Future work exploring alternative inputs to the selection mechanism \citep{Zattra2025ContextSelectiveSS} could help close the performance gap between SSMs and Transformers in query-first retrieval regimes.

\section{Acknowledgments} 
This work was supported in part by the Israel Science Foundation (grant no. 2045/21).

\subsection*{Limitations}

Our empirical evaluation focuses on Mamba-2 models, as open-source Mamba-3 models \citep{lahoti2026mamba3} are currently limited to small-scale research checkpoints released by the original authors. To the best of our knowledge, no larger-scale Mamba-3 checkpoints exist anywhere at present. These smaller parameter scales also lack the capacity to reliably execute the long-context key-value retrieval task examined in this work. Nonetheless, we conjecture that similar dynamics could persist in Mamba-3, as its gating design relies on comparable principles. We leave empirical validation on larger Mamba-3 checkpoints to subsequent work, once such models become available.

Beyond model scope, our analysis relies on a synthetic key-value retrieval task. While simpler than natural language, the controlled and predictable nature of the task was methodologically necessary, as it enabled the fine-grained analyses underlying our interpretability experiments. Crucially, key-value lookup represents a minimal requirement for long-context memory; that Mamba's selection mechanism fails even under these idealized conditions points to a fundamental architectural bottleneck rather than a task-specific artifact.

\subsection*{AI use statement}

In this work, we used generative AI tools to help develop theoretical models or conceptual frameworks, design or provide feedback on research methodology or experiments, and implement methods.
We have not used generative AI tools to generate synthetic data sets, clean and reformat dataset, assist with translation, support qualitative and thematic data analysis, interpret results, and propose or refine hypotheses.
Tasks such as formulate mathematical claims, provide critical ingredients for proving mathematical claims, assist in the writing of proofs---are not applicable to this work.
Additionally, we used generative AI tools to create or modify scientific figures or images, suggest experimental parameters, create or edit software code, brainstorm, edit a research paper to improve readability, and identify relevant literature.
We have reviewed all AI-assisted work and we take full responsibility for the final content of this work,
including text, claims or artifacts produced with the aid of generative AI.

\subsection*{Reproducibility statement}

All results presented in Sections \ref{sec:correlation}, \ref{sec:causality}, and \ref{sec:failure} are fully reproducible using the provided source code\footnote{Code is available at \url{https://github.com/schwartz-lab-NLP/beyond_state_capacity}}, which includes detailed environment and hardware specifications. Full implementation details are available in Appendix \ref{appendix:experiments_details}. The experimental datasets are generated automatically during execution, using hard-coded random seeds to ensure consistency.

\bibliography{references}
\bibliographystyle{iclr2027_conference}

\appendix
\crefalias{section}{appendix}
\section{Experimental Details}\label{appendix:experiments_details}

\begin{figure}[h]
\includegraphics[width=\linewidth]{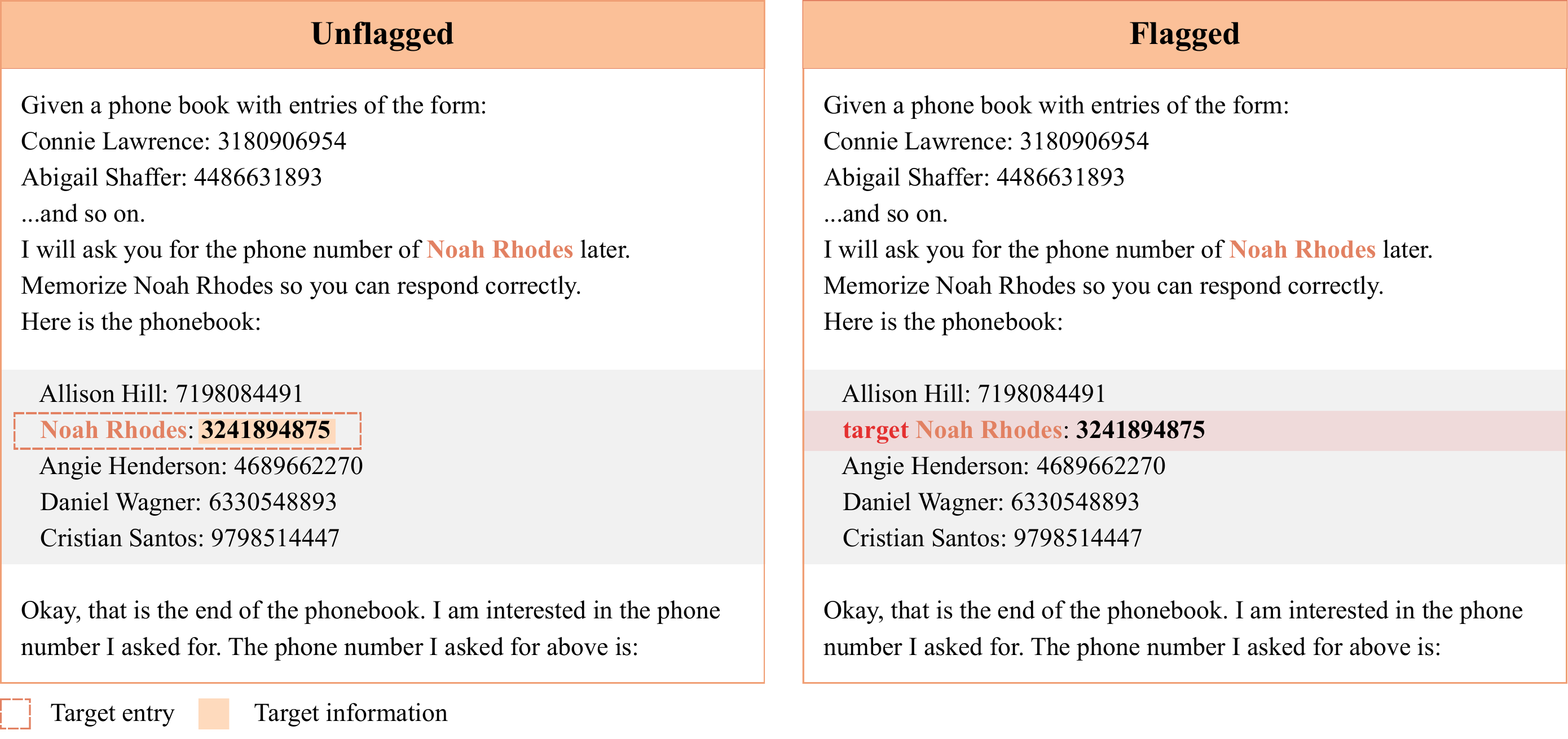}
\caption{A sample of \emph{unflagged} vs.~\emph{flagged} datasets. A single word addition, encoded to merely zero to two tokens, improves Mamba's accuracy substantially.}
\label{fig:prompts}
\end{figure}

\subsection{Datasets}
Each sample in the phonebook dataset (\cref{fig:prompts}) begins with a template instruction describing the task, which includes two randomly sampled demonstration entries separate from the actual phonebook. The prefix specifies the target person's name twice and prompts the model to memorize the corresponding phone number. The phonebook itself consists of a series of lines, where each line (referred to as an \emph{entry}) contains a name and a phone number separated by a colon and a space. The total number of entries defines the \emph{phonebook length}. 

Each phone number comprises 10 uniformly sampled random digits. Full names (first and last, no middle) are generated using the \texttt{Faker} python library to reflect realistic real-world name distributions. Following the phonebook, the prompt re-states the requirement for the previously requested target entry and prompts the model to complete the corresponding number. Samples are generated on-the-fly rather than drawn from a fixed static corpus, with random seed fixed to ensure exact experimental reproducibility (used $seed=42$).

The \emph{flagged} variant of the dataset differs solely by prepending the word \texttt{"target"} (followed by a whitespace) to the target entry within the phonebook. The \emph{profession-book} dataset---used exclusively in \cref{subs:state_probing}---uses a nearly identical template structure, but replaces phone numbers with professions sampled uniformly at random from a closed set of 20 categories (e.g., \texttt{"engineer"}, \texttt{"lawyer"}, \texttt{"doctor"}, \texttt{"teacher"}, \texttt{"architect"}).

\begin{figure}[h]
\includegraphics[width=\linewidth]{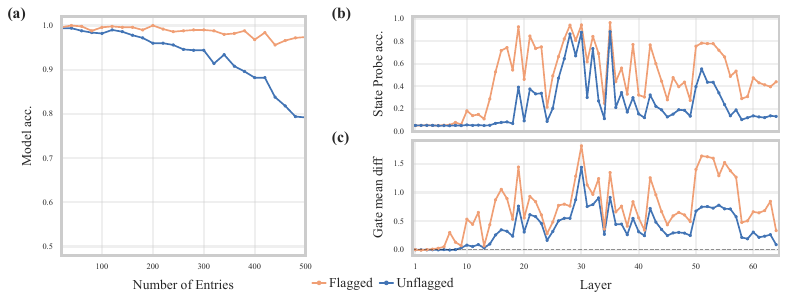}
\caption{\codestral exhibits trends similar to \nvidia when a local relevance flag is added to the target entry: (\textbf{a}) overall model accuracy improves, (\textbf{b}) the decodability of the target information in the recurrent state becomes stronger, and (\textbf{c}) selection gates better distinguish between target and distractor entries.}
\label{fig:codestral_corralation}
\end{figure}

\subsection{Models}

\paragraph{\nvidia} \citep{Waleffe2024empirical} Model weights are retrieved from the Hugging Face Hub under the repository \texttt{nvidia/mamba2-8b-3t-4k}. The weights are provided in a Megatron checkpoint format, which we remap to the official \texttt{mamba\_ssm} library structure. The model is a pure Mamba-2 base 8B-parameter language model pretrained on 3 trillion tokens. Architecturally, it comprises 56 layers with model dimension $d_{\text{model}} = 4096$, expansion factor $expand = 2$, head dimension $d_{\text{head}} = 64$ (yielding 128 total heads), state dimension $d_{\text{state}} = 128$, and $ngroups=8$.

\paragraph{\codestral} \citep{codestral2024blog} Model weights are retrieved from the Hugging Face Hub under the repository \texttt{mistralai/Mamba-Codestral-7B-v0.1}. The model is a 7.3B-parameter Mamba-2 model instruction-tuned for code generation. Architecturally, it comprises 64 layers with model dimension $d_{\text{model}} = 4096$, expansion factor $expand = 2$, head dimension $d_{\text{head}} = 64$ (yielding 128 total heads), state dimension $d_{\text{state}} = 128$, and $ngroups=8$. Presumably due to \codestral's instruction tuning, its accuracy is consistently higher than that of \nvidia, leading us to evaluate it on longer phonebooks to better amplify its tendencies. Unless stated otherwise, \nvidia is tested on phonebook length of 150 entries, and \codestral on phonebook length of 400 entries.

\paragraph{Tokenization} Both models employ SentencePiece tokenizers \citep{SentencePiece2018}. \codestral uses Mistral AI's native SentencePiece tokenizer with a 32,768-token vocabulary, whereas \nvidia uses NVIDIA's 256,000-token vocabulary SentencePiece model. Conveniently, both tokenizers process phonebook entries in a nearly identical way: name tokens (whose tokenization can differ) are followed by the separator sequence \texttt{":\_"}, 10 individual digit tokens, and a trailing newline token. This consistent tokenization enables clear alignment between target and distractor entries, as well as distinct separation between name and digit tokens. Prepending the word \texttt{"target"} to an entry in the \emph{flagged} dataset adds merely zero to two tokens to the prompt sequence. Additionally, all 20 professions in the profession-book dataset were selected such that each profession is tokenized into a single token across both tokenizers.

\paragraph{Hardware} All experiments were conducted on a single NVIDIA L40S GPU (48~GB VRAM), except for the state probing experiment (\cref{subs:state_probing}), which used two GPUs: one for model inference and state extraction, and the other for training the probes. Across both models, both datasets, and all experiments presented in this work, the total compute budget was roughly 1,000 single-GPU hours.

\subsection{Experiments}
\paragraph{Evaluation (\cref{subs:evaluation})} For each combination of dataset and phonebook length, model accuracy is evaluated over 500 dataset samples using a batch size of 1. Correctness is determined by generating 20 new tokens ($top\_k=1$), extracting a 10-digit sequence via regular expression, and comparing the extracted sequence against the ground truth. \nvidia is evaluated on phonebook lengths $L \in \{20, 40, \dots, 200\}$, while \codestral is evaluated on lengths $L \in \{20, 40, \dots, 500\}$.

\paragraph{State probing (\cref{subs:state_probing})} For each model and dataset, 15,000 samples are generated and partitioned into an 80/20 train/evaluation split for the probes. Models process inputs using a batch size of 12. State capture begins immediately after processing the target entry's profession token and continues after every subsequent profession token in the prompt, sampling up to 64 states per sample (evenly spaced along the sequence dimension). States are extracted separately for each model layer and flattened from their original 3D tensor shape $(n_{\text{heads}}, d_{\text{head}}, d_{\text{state}})$ into a 1D vector of dimension 1,048,576. All states associated with a given sample are assigned strictly to either the training set or the evaluation set to prevent data leakage. For each layer, we train an independent single-layer linear probe consisting of $\text{1,048,576} \times 20$ parameters. The probe weights are initialized in single precision (\texttt{fp32}) using Kaiming uniform initialization. Probes are trained using the Adam optimizer with a learning rate of $10^{-3}$, zero weight decay, and cross-entropy loss. To avoid writing hidden states to disk, model forward passes and probe updates are fully interleaved. Extracted states are fed into a class-balanced experience replay reservoir (128 slots per class), from which probes are trained using balanced mini-batches of size 64. Probe updates perform 16 training steps after every forward batch of 12 prompts. Updates begin only after two conditions hold: every class has at least one sample in the reservoir, and feature-wise mean and standard deviation have been frozen from a warmup of 4,096 captured states. Subsequent training and evaluation both use those frozen statistics to standardize states. Model evaluation follows the same online pipeline: hidden states are captured during forward passes and immediately evaluated by the probes, without the 64 states-per-sample limit and the replay buffer.

\paragraph{Gate distinction (\cref{subs:gate_distinction})} Each model is evaluated using 500 dataset samples. For every sample, entry, and digit token, we record the gate values across each (layer, head) pair independently to distinguish the target entry from distractor entries. For each layer, we average these gate values across all heads and samples within each category (target vs.\ distractor) and compute the difference $\text{target}_{\text{avg}} - \text{distractor}_{\text{avg}}$. Consequently, higher positive values indicate stronger model selectivity distinction between target and distractor entries. Throughout this paper, any experiment referencing \textit{gate} values uses $\Delta_t$, which is calculated by projecting the layer input $x_t$ into an intrinsic value (commonly denoted as $d_t$), adding a learned bias, and applying a Softplus activation. However, this specific experiment captures $d_t$ instead, as we found that averaging values after applying the activation function introduces artifacts into the results. This stems from the non-linearity of the Softplus activation, rather than the gates themselves: averaging post-activation values is not equivalent to averaging pre-activation values. Other experiments in this work produced equivalent results for $d_t$ and $\Delta_t$.

\paragraph{Patching and Scaling (\cref{sec:causality,subs:patching2})} For each model, 500 samples are generated and processed to evaluate performance and capture intermediate components for activation patching. Each sample is evaluated in both its \emph{flagged} and \emph{unflagged} variants to collect and align the corresponding internal activations. The patched components analyzed in this work are the selection gates $\Delta_t$ and the input representation to each layer $x_t$, which directly influences all downstream selection components (including the gates). Along the sequence dimension, activation patching is restricted exclusively to the 10 digit tokens corresponding to the target entry. Along the layer dimension, patching is applied cumulatively across layers---i.e., iteratively patching layer sets $\{1\}$, $\{1, 2\}, \dots, \{1, N\}$---with accuracy evaluated at each step. \cref{fig:failure} presents the results across all cumulative iterations, whereas \cref{fig:causality} shows the final iteration in which all $N$ layers are patched simultaneously. In the scaling experiment (\cref{fig:causality}(b)), instead of replacing $\Delta_t$ gates with those from another sample variant, we multiply the existing gate values by a scaling factor. Note that $\Delta_t$, which is obtained using a Softplus activation, is guaranteed to be strictly positive, ensuring that a scaling factor $> 1$ monotonically increases the gate magnitude while a scaling factor $< 1$ has the opposite effect.

\paragraph{Representation probing (\cref{subs:representation_probing})} For each model and dataset, 500 samples are generated and partitioned into an 80/20 train/evaluation split for probe training and testing. During model forward passes, we capture the hidden representations corresponding to the final digit token of each phone number entry in the phonebook. Representations are collected layer-wise, and an independent probe is trained for each layer. Each probe consists of $d_{\text{model}}$ weights (followed by a sigmoid activation) and outputs a single scalar probability; representations are classified as important using a decision threshold of 0.5. Probes are trained for 100 epochs over the complete set of extracted representations using the Adam optimizer with a batch size of 1024, a learning rate of $10^{-3}$, and zero weight decay. We optimize a binary cross-entropy loss weighted by the ratio of negative to positive samples to handle class imbalance. \cref{fig:failure} reports the $AUC$ score, which measures how well a classification model separates positive and negative classes.

\paragraph{Gate probing (\cref{subs:gate_probing})} For each model and dataset, 500 samples are generated and partitioned into an 80/20 train/evaluation split for probe training and testing. During model forward passes, we capture the gate vectors produced by the selection mechanism corresponding to the final digit token of each phone number entry in the phonebook. Probe training details are completely aligned with representation probing.

\end{document}